\documentclass[review]{elsarticle}

\usepackage{lineno,hyperref}
\modulolinenumbers[5]
\usepackage{amsmath}
\usepackage{amssymb}

\usepackage{xcolor}

\usepackage{multirow}
\newcommand{\revisions}[1]{{\color{black}{#1}}}
\journal{Journal of \LaTeX\ Templates}

\begin{document}

\begin{frontmatter}

\title{Real-time ground filtering algorithm of cloud points acquired using Terrestrial Laser Scanner (TLS)}


\author[mymainaddress]{Nelson Diaz\corref{mycorrespondingauthor}}
\cortext[mycorrespondingauthor]{Corresponding author}
\ead{ndiaz12@udi.edu.co}
\author[mysecondaryaddress]{Omar Gallo}
\author[mysecondaryaddress]{Jhon Caceres}
\author[mysecondaryaddress]{Hernan Porras}



\address[mymainaddress]{Department of Computer Science, Universidad de Investigación y Desarrollo, Bucaramanga, 680001 Colombia}
\address[mysecondaryaddress]{Department of Civil Engineering, Universidad Industrial de Santander, Bucaramanga, 680002 Colombia}

\begin{abstract}
3D modeling based on point clouds requires ground-filtering algorithms that separate ground from non-ground objects. This study presents two ground filtering algorithms. The first one is based on normal vectors. It has two variants depending on the procedure to compute the k-nearest neighbors. The second algorithm is based on transforming the cloud points into a voxel structure. To evaluate them, the two algorithms are compared according to their execution time, effectiveness and efficiency. Results show that the ground filtering algorithm based on the voxel structure is faster in terms of execution time, effectiveness, and efficiency than the normal vector ground filtering.
\end{abstract}

\begin{keyword}
Ground filter\sep normal vector\sep PCA\sep TLS\sep voxel.
\end{keyword}

\end{frontmatter}


\section{Introduction}
Since the development of LIDAR technology, it is easier to acquire the three-dimensional spatial information component in a wide range of scales \cite{diaz2014modelos}.  In the form of a point cloud, the way this technology captures data presents several challenges when processing this data. Among them, a significant challenge when processing point clouds in urban environments is filtering ground points, as this is essential for later segmentation and classification of objects.

Several methodologies have been proposed to perform ground filtering in Airborne laser scanning point cloud, such as, \cite{SITHOLE200485}, \cite{isprs-archives-XLIII-B1-2021-31-2021}. Weighted regression techniques are used in ground filtering in mobile laser scanner \cite{Nurunnabi2012DIAGNOSTICROBUSTSA}, \cite{7339614}, \cite{8681640}. And ground filtering techniques in terrestrial laser scanner usually are based on normal vectors.  This approach computes the corresponding normal vector and surface vector from the neighborhood of a point \cite{isprs-archives-XXXVIII-5-W12-109-2011}. Once they are calculated, the point is considered as belonging to the ground if the angle between these two vectors is close to 90 degrees. Some points on trees and buildings may be erroneously classified as ground; hence, these points are eliminated by adjusting to a plane.  To do this, a best-fitting plane to the points classified as the ground is calculated.  Then, the distance from each point to the plane is measured, and the point with the minimum normal distance to the plane is determined. This point is called $Z_0$. After, an elevation threshold is set according to $Z_0$, and the points above and below the threshold are removed.  This leads to ground filtering that leaves only the curb's limits, road lanes, and sidewalks.

Other methods to filter ground are based on structures that divide the point cloud into small cubic sections called voxels \cite{6466890}. The method identify the lowest cubic section that is full, which means a section that contains at least one point in each of the vertical columns of voxels. Voxels with these characteristics are considered to contain ground points. Subsequently, from the points identified as ground, the plane that best fits these points is calculated. The nearest points to the plane will be classified as ground points. It should be noted that the use of the voxel concept reduces the computational complexity of the problem since it does not work directly on the points but on their aggregation as a voxel. This approach has been successfully used in real-time curb detection applications for mobile laser scanners \cite{6466890}. Approaches involving voxels have also been used for point cloud segmentation and classification \cite{Douillard2014}. 

In \cite{isprs-archives-XXXIX-B5-187-2012} the point cloud is divided into vertical columns, and each vertical column is divided into cubes.  For each point, the region of the vertical elevation histogram in which it is located is calculated.  The vertical elevation histogram shows the number of points in each cube. The lowest cube in the vertical elevation histogram is then selected as ground.

In order to improve the segmentation of ground points, the point cloud must be organized before processing.  This organization reduces the computation time of certain operations, such as nearest neighbors' calculation, easing the segmentation process.  There are different algorithms for structuring a point cloud.  Among them, we can mention those based on the creation of a kd-tree \cite{10.1145/361002.361007} and an octree \cite{MEAGHER1982129}.  These algorithms organize the points in a tree-like data structure, which facilitates finding nearby points quickly by searching through the created tree.  However, the search of nearby points is possible only after the computation of the tree. Also, the organization of point clouds with these structures is performed on the totality of available data. However, the reorganizing of point cloud is inconvenient at the time of processing because the spatial characteristics of closer points are alike, making it unnecessary to study distant points (the proximity is characterized according to the neighborhood concept between points in space, see proposed methodology).  Therefore, it is more convenient to structure the point cloud into smaller sections. From now on, these sections will be referred to as partitions and voxels.

This article presents two ground-filtering algorithms. The first one uses normal vectors and has two versions that vary concerning on calculating k-nearest neighbors. K-nearest neighbors refer to the nearest k points to a $\mathbf{p}_i$ point. The first version uses a general-purpose library to calculate k-neighbors called KNN CUDA; the second one uses a point cloud structuring algorithm called kd-tree from the VTK® library.  VTK is a C++ library for image processing and graphic computing.  In particular, VTK allows structuring point clouds using Kd-trees.  The second ground filtering algorithm presented is based on the point cloud's structuring into voxels, which groups nearby points, enabling operations to be performed on the voxels instead of the points.  The results show that the voxel-based ground filtering algorithm is faster in terms of execution time than the normal vector-based ground filtering algorithm.

This work is organized as follows: Section 2 briefly explains the characteristics of the data used to test the algorithms, including the specifications of the hardware used to implement the proposed algorithms and the detailed description of the ground filtering algorithms developed, i.e., one based on normal vectors and the other based on voxel structuring.  In Section 3, the experimentation and discussion of results can be found. Finally, Section 4 presents the conclusions of this paper.

\section{Methodology}
The experimental scheme used for the present study is detailed below.  Section 2.1 explains the data used.  Section 2.2 details the hardware used to run the algorithms.  Finally, section 2.3 describes the two ground filtering algorithms: the first based on normal vectors, offering two versions that differ in the way they calculate the neighbors of a point, and the second one, based on voxel structuring.

\subsection{Data}

For the present study, data collected on the Carrera 27 from the 14\textsuperscript{th} to 18\textsuperscript{th} street corners in the city of Bucaramanga, Colombia, were used.  The test point cloud was captured using the Riegl VZ-400 terrestrial laser scanner.  In total, 11-point clouds were used. The number of points in each cloud is between 1047348 and 5523518.

\subsection{Hardware and libraries used}
The proposed algorithm was tested on a machine with the following specifications: operating system Windows 10, an intel(R) Core(TM) i7-10750H 2.60 GHz, and a 6 GB NVIDIA GeForce RTX 2060 card. The algorithm was written in C/C++ language, and the compilers used were gcc and g++. In addition, the KNN CUDA and kd-tree VTK® algorithms were used to calculate the k-nearest neighbors. The source code of the proposed approach is available in the following repository\footnote{The source code are available in the following \href{https://github.com/nelson10/Point-cloud-filtering.git}{GitHub repository.}}.


\subsection{Filtering algorithms}
 In order to establish the nomenclature to be used throughout this paper, the following definitions are proposed.  Let $\mathbf{P} \in \mathbb{R}^{M \times N}$ be the matrix that represents a 3D point cloud with $N$ points. In more detail, each column of $\mathbf{P}$ is a 3D vector,  such that, $\mathbf{P}=[\mathbf{p}_1,\dots,\mathbf{p}_i,\dots,\mathbf{p}_N]$, where $\mathbf{p}_i \in \mathbb{R}^M$ is the $i\textsuperscript{th}$ point in the point cloud denoted by $\mathbf{p}_i=[p_{(x,i)},p_{(y,i)},p_{(z,i)}]^T$, and $i$ is the index of the columns in $\mathbf{P}$. The nearest points to point $\mathbf{p}_j$ are called the neighbors and are grouped according to the neighborhood relationship. Given the query point $\mathbf{p}_i$ the resulting $K$-nearest neighborhood matrix is given by $\mathbf{Q}_i \in \mathbb{R}^{M \times K}$, where each column is a 3D vector, such that, $\mathbf{Q}_i =[\mathbf{q}_1,\dots,\mathbf{q}_k,\dots,\mathbf{q}_K]$, with $K$ being the number of neighbors of point $\mathbf{p}_i$. In the present research, two methodologies for ground filtering were studied to evaluate which of them presented the best performance and efficiency. The first one is based on the calculation of normal vectors for each point, and the second one is based on the voxel construction technique.

\subsubsection{Normal vector calculation technique}
Figure 1. depicts the flowchart of the proposed approach. Six stages compose the algorithm, standardization, construct of 2D voxel grid, search of the k-nearest neighbors, principal component analysis, adjusting of ground points to a plane, and ground selection. In the following, each stage is explained.

\textit{Standardization:} Before performing the calculations for the ground filtering algorithm, it is convenient to standardize the data. This is due in principle to two reasons: the overflow of memory by the calculations made on floating data types and eliminating the outliers. In the first one, the coordinates $(x,y,z)$ of the point clouds are floating data types. For this reason, the calculations performed on the video card may overflow the memory as the capacity supported for this type of data is exceeded. This results in a loss of calculation accuracy.  On the other hand, the point cloud includes some atypical values that deviate from the main group of points in the scene. These values are called outliers, and their presence increases the extreme values in the coordinates $(x,y,z)$.  The standardization process was carried out by applying the logistics function (Verhulst, 1845) at each coordinate $(x,y,z)$  for all points in the cloud. An example of standardization for the $x$ coordinate is shown in equation 1, where $\hat{x}_i$ is the logistics function of the  $i$\textsuperscript{th} point of the $x$ dimension. In this, $\tau_i$ is dependent on four parameters: $x_i$ is the $x$ component of the $i$\textsuperscript{th} point; the variables $\bar{p}_x = \frac{1}{N} \sum_{i=1}^N p_{(x,i)}$ and $\sigma_x= \frac{1}{N}(p_{(x,i)}-\bar{p}_x)^2$ are the mean and standard deviation of the $x$ dimension; and the parameter $r$ indicates how many standard deviations are considered.  In the present case, $r=1$. The standardized points are within the interval $[0-1]$.
\begin{figure}[!htb]
	\centering
	\includegraphics[scale=0.5]{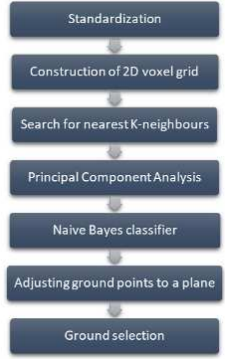}
	\caption{Flowchart of ground filtering algorithm based on normal vector calculation.}
\end{figure}

\begin{equation}
\hat{p}_{(x,i)} = \frac{1}{1 + \exp{(-\tau_i})}, \hspace{0.3cm}\text{  where  } \hspace{0.3cm}\tau_i = \frac{p_{(x,i)}-\bar{p}_x}{r\sigma_x}.
\end{equation}

\begin{figure}
\begin{center}
\begin{tabular}{c c c}
    \hspace{-12pt}
    \rotatebox[origin=c]{90}{\parbox{0.20\linewidth}{\centering \footnotesize Pre-processing}}
    &
     \hspace{-15pt}
     \begin{minipage}{.435\linewidth}
     \includegraphics[width=\linewidth]{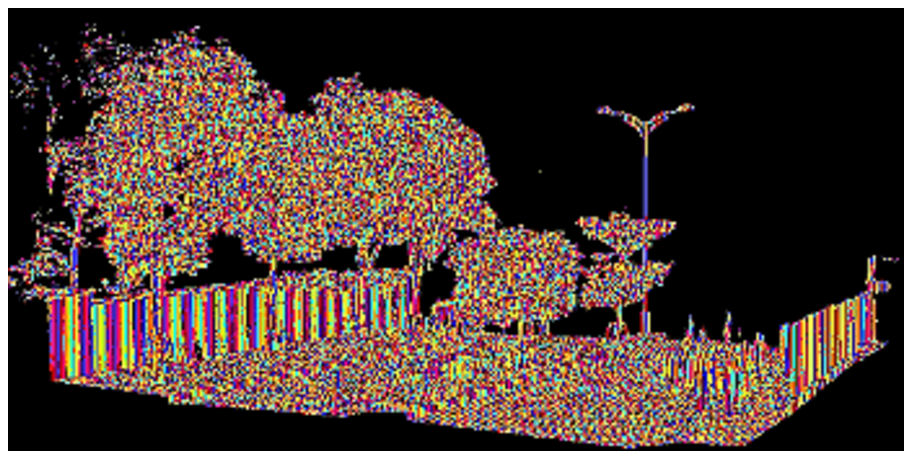}
     \end{minipage}
     & 
     \hspace{-15pt}
     \begin{minipage}{.435\linewidth}
     \includegraphics[width=\linewidth]{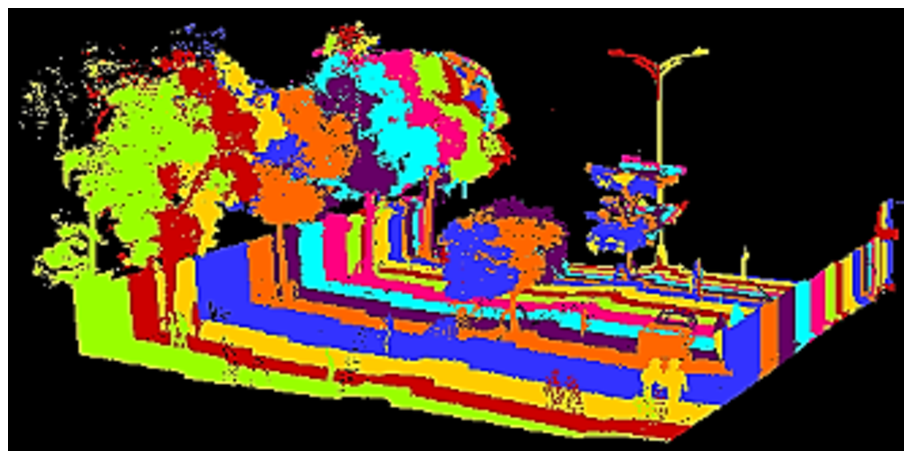}
     \end{minipage}
    \\
    &
     \hspace{-15pt}
     \scriptsize (a) 
     &
     \hspace{-15pt}
     \scriptsize (b) 
    \\
    \hspace{-12pt}
    \rotatebox[origin=c]{90}{\parbox{0.20\linewidth}{\centering \footnotesize Ground filtering}}
    &
     \hspace{-15pt}
     \begin{minipage}{.435\linewidth}
     \includegraphics[width=\linewidth]{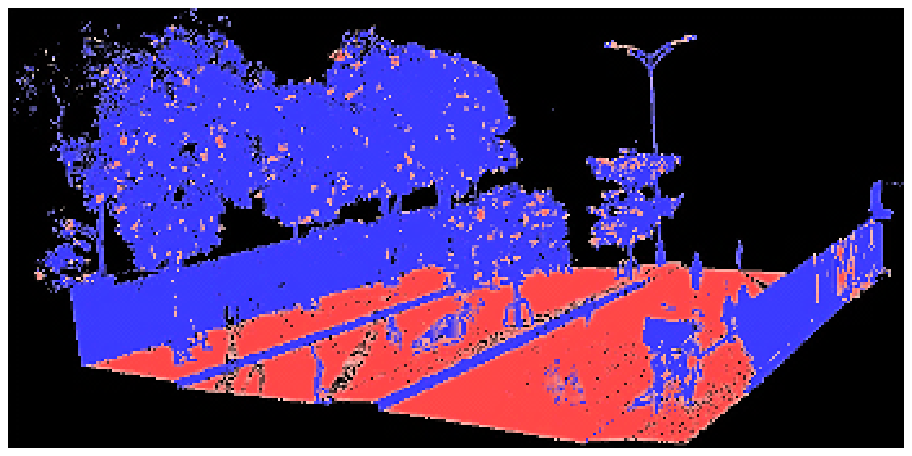}
     \end{minipage}
     & 
     \hspace{-15pt}
     \begin{minipage}{.435\linewidth}
     \includegraphics[width=\linewidth]{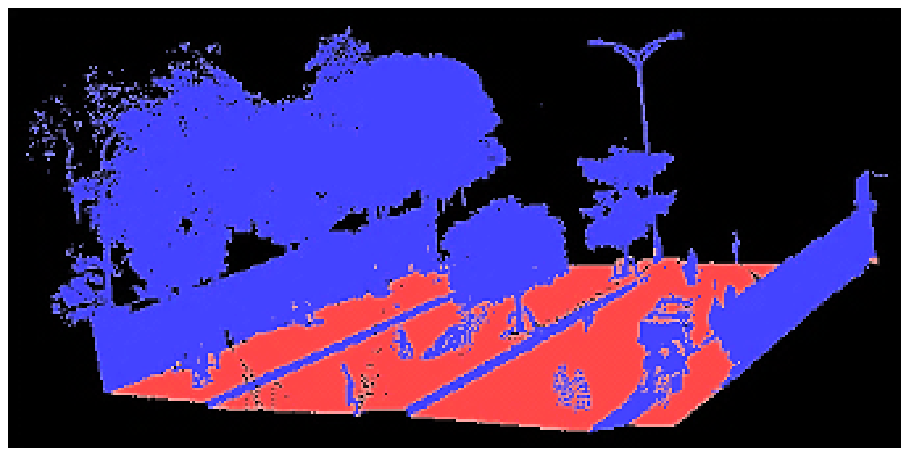}
     \end{minipage}
     \\
     &
     \hspace{-15pt}
     \scriptsize (c) 
     &
     \hspace{-15pt}
     \scriptsize (d) 
\end{tabular}
\end{center}
    \vspace{-10pt}
    \caption{\footnotesize{(a) 2D Grid division, (b) Segments with 65,536 points, (c) misclassified ground filtration, (d) Ground filtering after the stages of the Näive Bayes classifier and adjustment of points to the plane.}} \label{fig:2}
    \vspace{-10pt}
\end{figure}

\textit{Grid Construction}: To avoid the assignment of points that are far from each other to the same neighborhood, a mechanism was proposed to organize the point cloud spatially and thus achieve better performance in searches. This organization consists of building a 2D grid using the $x$ and $y$ axes of the point cloud and then gridding the space at 0.1-meter intervals for each axis. These values were determined by experimentation. It is common in the literature for this parameter to be established experimentally. For instance, \cite{6466890} have defined $0.2$ meters as the voxel dimension.  This grid has no limit on the $z$ coordinate, so each cell contains all the points within its $x$ and $y$ cell ranges, forming a type of vertical columns of points \cite{6466890}, as shown in figure 2(a). These vertical columns are grouped sequentially to form segments of $65,536$ points as shown in figure 2(b). The segments have this size since it is the maximum number of points that can be processed with KNN CUDA, according to tests shown in \cite{4563100}, \cite{5654017}. \revisions{The K-nearest neighbors in the edge of the segment with 65536 are calculated using only points in the segment. This approach does not consider neighbors of adjacent segments because this increases the computational time.}

\textit{Normal vector calculation}: The normal vector of each cloud point is calculated by using an algorithm that consists of two stages: the first one searches for the $K$-nearest neighbors (KNN), and the second one uses the analysis of the main components or Principal Component Analysis (PCA).

\textit{Search for K-NN}: During this stage, the computing K-NN was conducted in two ways: by structuring the point cloud in the form of a kd-tree via the VTK® library \cite{567752}, \cite{865875}, and using the parallel programming paradigm through the KNN CUDA library to calculate $K$-neighbors for each of the query $\mathbf{p}_i$ points \cite{4563100}, \cite{5654017}. The video card processors of the machine used in this study were also used. The number of neighbors is set to $K = 50$. The comparison of the time used under both approaches is shown in section 3, corresponding to the results.

\textit{Principal Component Analysis:} The principal component analysis (PCA) is computed from points in the $K$-neighborhood $\mathbf{Q}_i$ of a giving query point $\mathbf{p}_i$, where $i \in [1,N]$ is any point in the point cloud. Specifically, the first step to obtain the PCA is computing the covariance matrix $\mathbf{S}_{\mathbf{q}} \in \mathbb{R}^{M \times M}$ using equation the following equation

\begin{equation}
   \mathbf{S}_{\mathbf{q}} = \frac{1}{K} \sum_{k=1}^K \mathbf{q}_k \mathbf{q}^T_k - \mathbf{m}_\mathbf{q} \mathbf{m}^T_\mathbf{q}  = (\mathbf{q}_k - \mathbf{m}_\mathbf{q})(\mathbf{q}_k - \mathbf{m}_\mathbf{q})^T,  
\end{equation}

\noindent where $\mathbf{m}_\mathbf{q}= \frac{1}{K} \sum_{k=1}^{K}\mathbf{q}_k$ is the mean vector in a $K$-neighborhood. Then, finding a set of $M$ orthonormal eigenvectors $\mathbf{v}_j$ for $j=1,\dots,M$ is always possible because $\mathbf{S}_{\mathbf{q}}$ is real and symmetric matrix. Solving the following equation $\mathbf{S}_{\mathbf{q}} \mathbf{v}_j = \lambda_j \mathbf{v}_j$ produces eigenvalues $\boldsymbol{\lambda} = [\lambda_1,\lambda_2,\lambda_3]^T$, and eigenvectors $\mathbf{v}_j$ for the points in the neighborhood. After that, the eigenvalues are normalized according to

\begin{equation}
    Nor(\lambda_j) = \frac{\lambda_j}{\lambda_1 + \lambda_2 +\lambda_3}.
\end{equation}

Then, the normal vector to the surface corresponds to the eigenvector of the smallest eigenvalues. Furthermore, three descriptors are defined to characterize the type of shape of each $\mathbf{Q}_i$ neighborhood: linear, flat, and volumetric \cite{isprs-annals-I-3-245-2012}. Finally, each point is classified as a possible ground according to the angle between the normalized eigenvalues $Nor(\lambda_j)$  and the normal vector to the surface. The angle should be found at the threshold [80,100] degrees according to \cite{isprs-archives-XXXVIII-5-W12-109-2011}, and the geometric descriptor should be flat. If these two characteristics are met, the $\mathbf{p}_i$ point is classified as ground. Two matrices result once each point is classified. Specifically, the matrix $\mathbf{G} \in \mathbb{R}^{M \times U}$ represent the $U$ point classified as ground. The matrix $\mathbf{F} \in \mathbb{R}^{M \times V}$ denotes the $V$ points classified as non-ground, where $V = N-U$.

\textit{Naïve Bayes Classifier:} From the previous stage, most of the points that were labeled as ground belongs to it. However, some points belonging to other objects such as trees or buildings also met the condition of the angle between their normalized eigenvalues and the normal vector and were therefore labeled as ground.
In order to improve the previous classification, an algorithm based on the probability of a point belonging to one of the two existing classes (ground and non-ground) called the Naïve Bayes classifier was used. This algorithm performs a neighborhood analysis of each point based on a sample and the total number of points.
For the case study, the a priori probability of points on the ground $P(U)$  is defined as the number of points on the ground $U$ over the total number of points $N$, see Equation (4). The a priori probability of points that are not ground $P(V)$ is defined as the number of points that are not ground $V$ over the total number of points $N$, see Equation (5).
\begin{equation}
    P(U) = \frac{U}{N}
\end{equation}
\begin{equation}
    P(V) = \frac{V}{N}.
\end{equation}
For each  $\mathbf{p}_{i}$ query point, its neighborhood $\mathbf{Q}_i$  and its probability of belonging to $\mathbf{Q}_i$  are calculated using equations (6) and (7), where $P(U| \mathbf{Q}_i)$ and $P(V |\mathbf{Q}_i)$ are the probabilities that the points in the neighborhood have of belonging to one of the two classes. Where the number of ground point in $\mathbf{Q}_i$ is denoted by $U_{\mathbf{Q}_i}$ and the number non-ground points in $\mathbf{Q}_i$ is denoted by $V_{\mathbf{Q}_i}$, and $K$ is the number of points in the neighborhood of $\mathbf{p}_i$.
\begin{equation}
    P(U_\mathbf{Q_i}|\mathbf{Q}_i )=\frac{U_{\mathbf{Q}_i}}{K},
\end{equation}
\begin{equation}
    P(V_\mathbf{Q_i}|\mathbf{Q}_i)=\frac{V_{\mathbf{Q}_i}}{K}.
\end{equation}
Finally, the a posteriori probability is calculated for the points that belong and do not belong to the ground, see equation (8) and (9).
\begin{equation}
     Pp(U) = P(U)P(U_{\mathbf{Q}_i}|\mathbf{Q}_i),
\end{equation}
\begin{equation}
    Pp(V)=P(V)P(V_{\mathbf{Q}_i}|\mathbf{Q}_i). 
\end{equation}
 
The point is classified as ground if $Pp(U)> Pp(V)$ or as not ground if $Pp(U)< Pp(V)$. The application of the Naïve Bayes classifier substantially improved the classification of the points. However, some points remained misclassified, as shown in Figure 2(c). For this reason, the strategy of adjusting the points that belong to the ground to a plane was used in order to eliminate the points that remain misclassified.

\textit{Adjusting ground points to a plane:} The points that were classified as ground were adjusted to a plane defined by $\alpha x+\beta y+ \gamma z+\delta=0$. The plane is obtained using the RANSAC algorithm that calculates the coefficients $\alpha,\beta,\gamma,\delta$, then the distance of each point $\mathbf{p}_i$ to the plane is calculated
\begin{equation}
    d_i=\frac{\alpha x_j+\beta y_j+\gamma z_j+\delta}{\sqrt{\alpha^2+\beta^2+\gamma^2}}.
\end{equation}
Based on the distances $d_i$, the mean distance $\mu$ and its standard deviation $\sigma$ are calculated. The points are then classified as ground if they meet two criteria: the first is that their distance $d_i$ to the plane is within the distance threshold of 10 cm, a value set based on experimentation. The second is that $d_i$ is also within the value of $\mu \pm \sigma$. The result is shown in Figure 2(d).



\subsubsection{Voxel-based ground filtering algorithm}

Another approach studied for ground filtering is the voxel-based filtering algorithm. Figure 3 shows the flowchart for the developed algorithm.

\begin{figure}[!htb]
	\centering
	\includegraphics[scale=0.5]{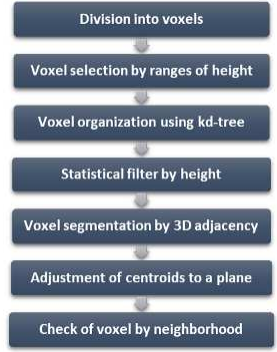}
	\caption{Flowchart of voxel-based ground filtering algorithm.}
\end{figure}

\textit{Division into voxels:} The voxel-based division is a 3D spatial segmentation that splits the 3D space and assigns each point to a segment, based on the three coordinates $(x,y,z)$ of each point.  This division of space into cubes called voxels transforms the point cloud into a three-dimensional mesh, reducing computational complexity in nearest neighbors' calculation.  The size of the voxel is $10$ cm. This size was determined experimentally and based on the work of \cite{6466890}. It should be noted that the voxel division is similar to the grid construction of the ground filtering algorithm based on normal vectors, but with the particularity of dividing also the $z$ dimension. The result can be seen in Fig. 4(a).

\textit{Voxel selection by ranges of height:} The criterion used to detect ground points using the voxel division is based on the fact that ground sections are nearly flat. Therefore, in each voxel, the $z$ coordinate range is calculated as follows: $range = max_z–min_z$. This is done to quantify how flat the point cloud section contained in each voxel is.  If the range is below 0.04 cm, the voxel is considered a horizontal surface. This parameter was defined by experimentation results are shown in Figure 4(b).

\textit{Voxel organization using kd-trees:} To structure the point cloud, a kd-tree is created using the centroid of each voxel. This organization reduces the complexity of searching for nearest neighbors. The centroid is calculated as shown in equation 11.  Let $\mathbf{V}_\ell \in \mathbb{R}^{M \times L}$  be the matrix that groups all the $L$ points within a $\ell\textsuperscript{th}$ voxel.  The  $x$-coordinate of the centroid for the points within the voxel is calculated by

\begin{equation}
    c_{x,\ell} = \frac{max(V_{(x,l)})+min(V_{(x,l)})}{2} \hspace{0.5cm} \ell=(1,\dots,T),
\end{equation}
\noindent where $T$ denotes the total number of voxels. A similar calculation is performed for the $y$  and $z$ coordinates, defining the centroid $\mathbf{c}_\ell=[c_{(x,\ell)},c_{(y,\ell)},c_{(z,\ell)}]^T$  . This point is representative of all points within the voxel.  A kd-tree is created with all centroids, which makes it possible to calculate neighborhoods among voxels.

\begin{figure}
\begin{center}
\begin{tabular}{c c c}
    \hspace{-12pt}
    \rotatebox[origin=c]{90}{\parbox{0.20\linewidth}{\centering \footnotesize Pre-processing}}
    &
     \hspace{-15pt}
     \begin{minipage}{.435\linewidth}
     \includegraphics[width=\linewidth]{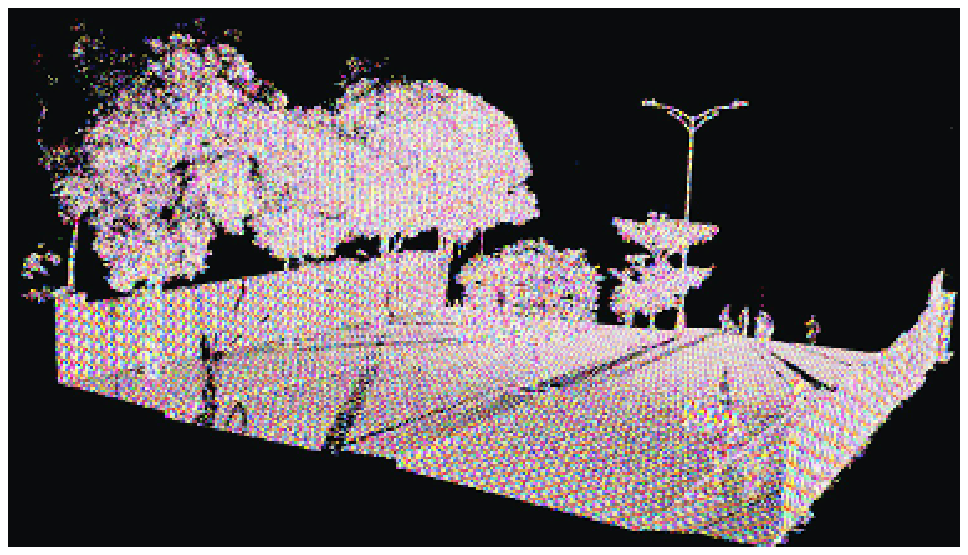}
     \end{minipage}
     & 
     \hspace{-15pt}
     \begin{minipage}{.435\linewidth}
     \includegraphics[width=\linewidth]{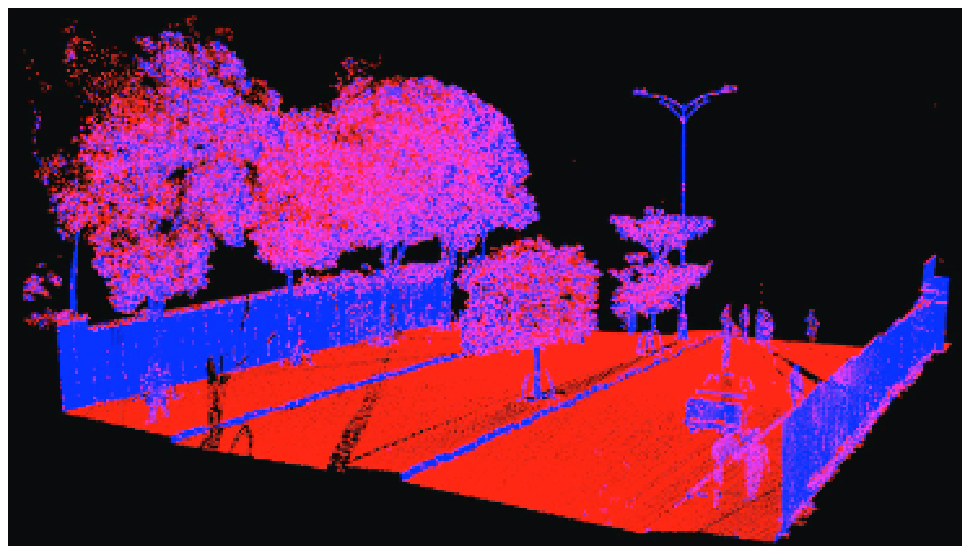}
     \end{minipage}
    \\
    &
     \hspace{-15pt}
     \scriptsize (a) 
     &
     \hspace{-15pt}
     \scriptsize (b) 
    \\
    \hspace{-12pt}
    \rotatebox[origin=c]{90}{\parbox{0.20\linewidth}{\centering \footnotesize Pre-processing}}
    &
     \hspace{-15pt}
     \begin{minipage}{.435\linewidth}
     \includegraphics[width=\linewidth]{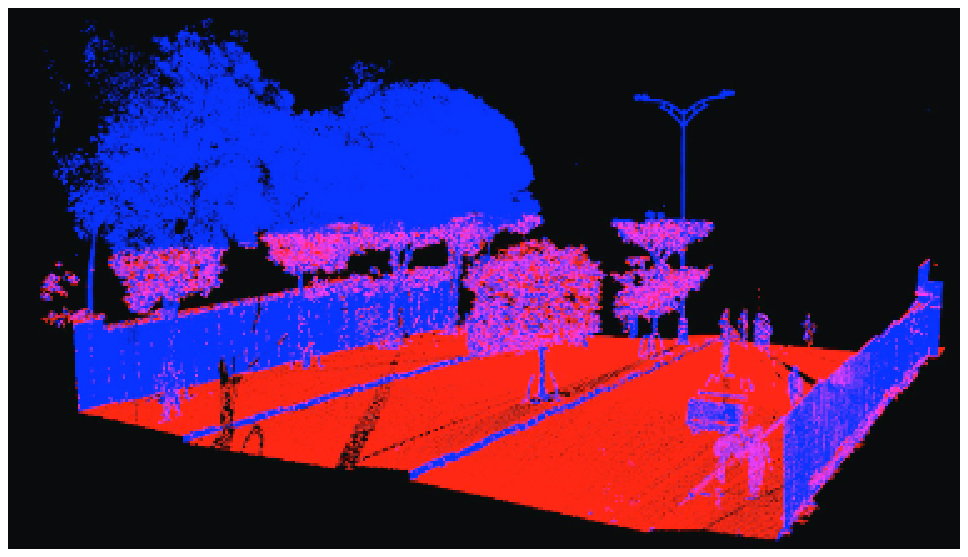}
     \end{minipage}
     & 
     \hspace{-15pt}
     \begin{minipage}{.435\linewidth}
     \includegraphics[width=\linewidth]{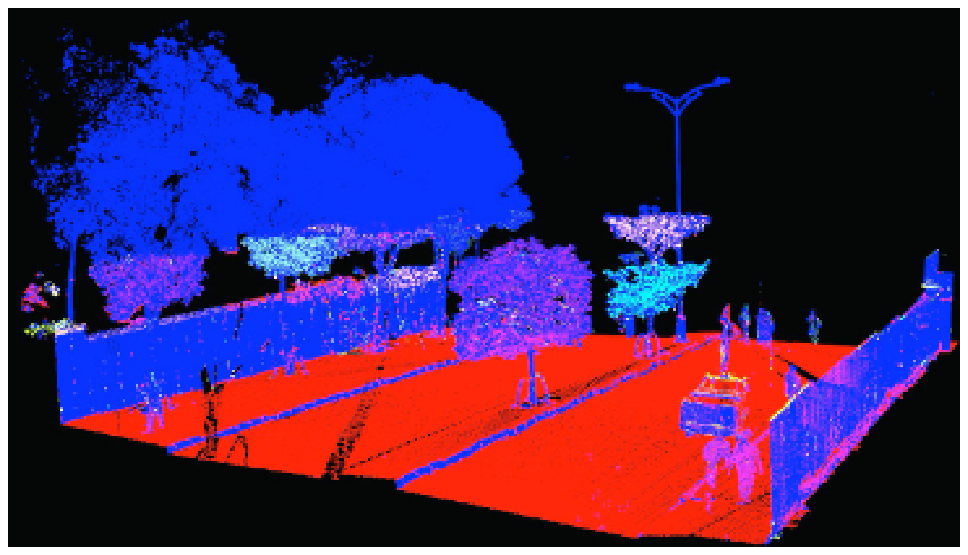}
     \end{minipage}
     \\
     &
     \hspace{-15pt}
     \scriptsize (c) 
     &
     \hspace{-15pt}
     \scriptsize (d) 
     \\
    \hspace{-12pt}
    \rotatebox[origin=c]{90}{\parbox{0.20\linewidth}{\centering \footnotesize Ground filtering}}
    &
   \multicolumn{2}{c}{
        \hspace{-15pt}
     \begin{minipage}{.435\linewidth}
     \includegraphics[width=\linewidth]{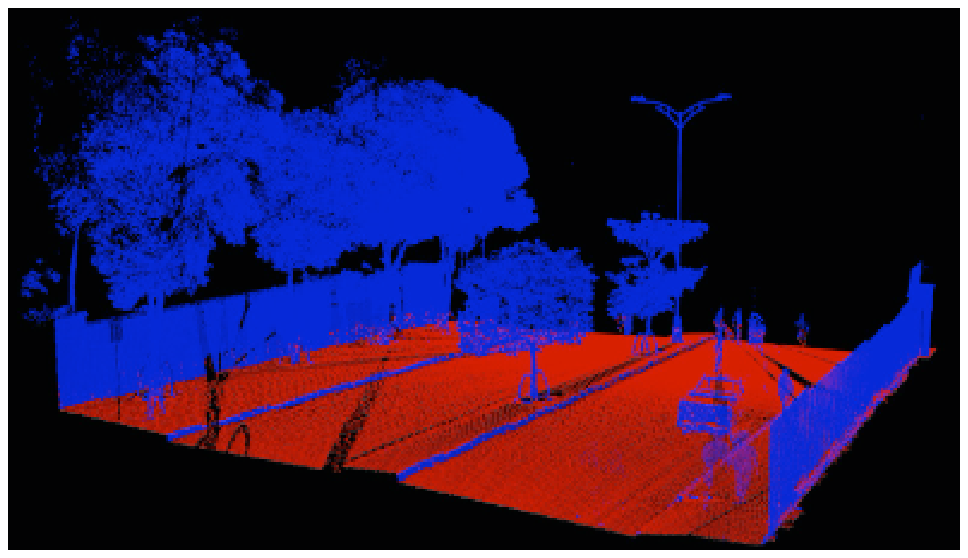}
     \end{minipage}
   }

     \\
     &
     \multicolumn{2}{c}{
     \hspace{-15pt}
     \scriptsize (e) 
     }

    \\
\end{tabular}
\end{center}
    \vspace{-10pt}
    \caption{\footnotesize{The figure depicts the steps of the voxel algorithm, which has 5 parts. (a) Division into voxels, (b) Voxel selection by ranges of height, (c) Statistical filter by height, (d) voxel segmentation by 3D adjacency, (e) Check of voxel by neighborhood.}} \label{fig:4}
    \vspace{-10pt}
\end{figure}

\textit{Statistical filter by height:} In order to reduce the number of voxels that are the farthest from the ground in a preliminary way, a statistical filter by height is applied.  More specifically, for voxels with a flat surface, i.e., with a range of 0.04 cm, the $z$-coordinate of their respective centroid was averaged. From this average, the horizontal surface threshold, shown in equation 12, was calculated. The threshold is equal to the average of the horizontal surfaces plus one meter.
\begin{equation}
    threshold_{sh}= average_{sh}+ 1.
\end{equation}
Finally, the voxels below $ threshold_{sh}$ are selected (See Fig. 4 (c)), thus discarding the voxels farthest from the ground.

\textit{Voxel segmentation by 3D adjacency:}  The voxels resulting from the previous step are arranged into segments.  These segments are constructed using a seed growth algorithm \cite{pu2006automatic}, which calculates each voxel's neighborhoods using adjacency 26. Adjacency 26 is defined as the neighborhood around a voxel \cite{Douillard2014}.  See figure 4 (d), showing groups of voxels.

\textit{Adjustment of the centroids to a plane:} After the voxel segmentation, the points within the largest voxel segment are adjusted to a plane.  This plane is calculated using the centroids of the largest segment.  The distance from each centroid to the plane is then calculated according to Equation (10). The points within the voxels that meet the criterion of adjustment of points from the ground to a plane on section 2.3.1 and then have at least 2 points are selected.

\textit{Check of voxel by neighborhood:} To verify the classification of points within the voxel as ground and not ground, the number of ground and non-ground voxels in adjacency 26 are counted.  The points within the considered voxel are reclassified according to the most significant number of neighboring voxels. The filtering result is shown in Figure 4(e).

\section{Experimental Results}

This section discusses the results of the ground filtering algorithms in terms of effectiveness and efficiency. The analysis includes a comparison between the execution time of the ground filtering algorithm based on normal vectors varying the function that calculates the nearest neighbors since that is the algorithm's bottleneck. 

\begin{figure}[h!]
    \begin{center}
    \begin{tabular}{c c c c}
    \hspace{-10pt}
    \includegraphics[width=.49\linewidth]{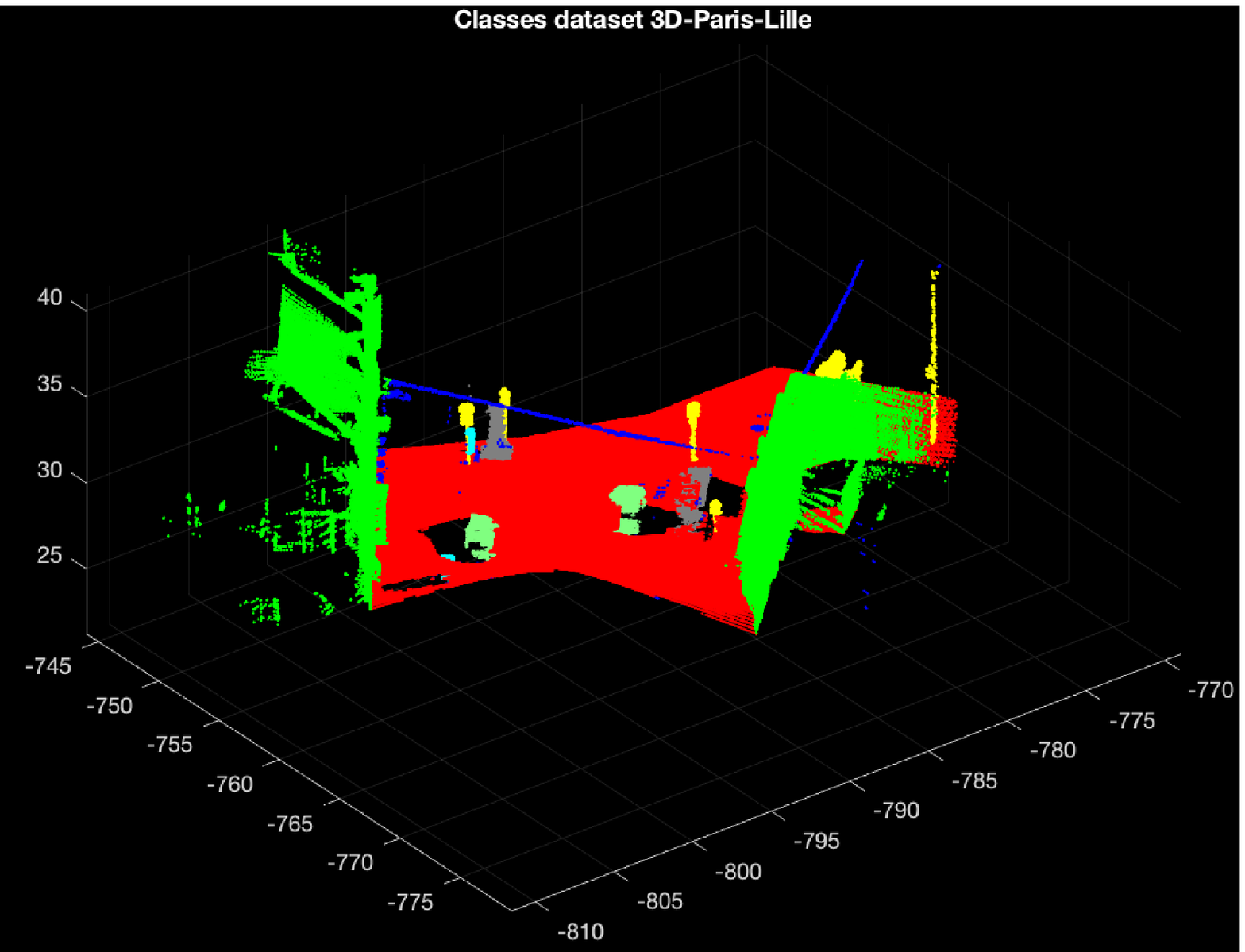}
    &
    \hspace{-12pt}
    \includegraphics[width=.49\linewidth]{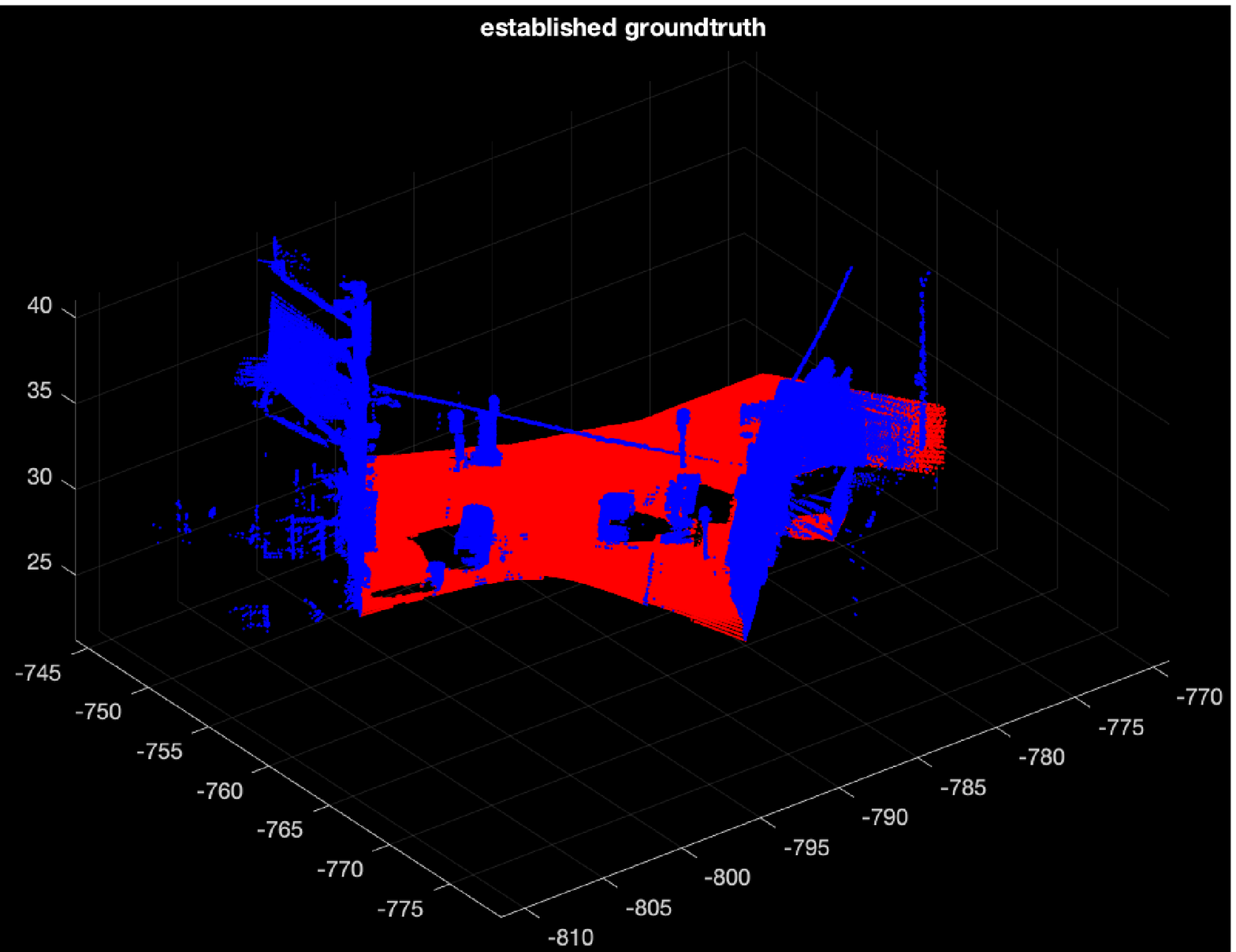}
    \\
    \footnotesize (a) 
    &
    \footnotesize (b)
    \end{tabular}\vspace{-8pt}
    \end{center}
    \caption{\revisions{3D Paris-Lille dataset. (a) 3D Paris-Lille classes, (b) established groundtruth map.}\vspace{-15pt}}
    \label{fig:paris-lille_image_gt}
\end{figure}

\revisions{Furthermore, simulations with a benchmark dataset are included to evaluate the proposed normal-vector and voxel-based ground filtering approaches. The selected benchmark is the Paris-Lille-3D dataset \cite{roynard:hal-01695873}, which contains ten classes, i.e., class 0 unclassified, class 1 ground, class 2 building, class 3 pole - road sign - traffic light, class 4 bollard - small pole, class 5 trash can, class 6 barrier, class 7 pedestrian, class 8 car, class 9 natural - vegetation.  Specifically, Fig. \ref{fig:paris-lille_image_gt}(a) depicts the different classes of one part of the scan of the Paris-Lille-3D dataset.  For the present study, it is preserved the ground class, and the other classes are set as non-ground, as it is shown in Fig. \ref{fig:paris-lille_image_gt}(b). Four labeled scans of the Paris-Lille-3D dataset are utilized to evaluate the performance of normal-vector and voxel-based ground filtering. In addition, five metrics are included to assess the performance of the ground filtering approaches.}

\subsection{Accuracy Evaluation Metrics}

\revisions{To test the quality performance of the proposed methods, standard metrics for semantic segmentation and binary classification tasks are used. Let TP, FP, FN, and TN represent true positives, false positives, false negatives, and true negatives, respectively. Then, the accuracy is given by}
        
        \begin{equation}
            Accuracy =  \frac{(TP + TN)}{(TP+FP+TN+FN)}.
        \end{equation}
        
\revisions{The precision/correctness, recall/completeness are computed as follows:}
        
        \begin{equation}
            Precision =  \frac{TP}{(TP + FP)},
        \end{equation}
        
\revisions{ \noindent where the precision is the ratio of the true positives over the extracted ground points, and the recall is the ratio of the true positives over the labeled ground-truth ground points. The higher the value of the metrics, the better the performance of the methods.}
        \begin{equation}
            Recall  =  \frac{TP}{(TP + FN)}.
        \end{equation}
        
\revisions{Furthermore, it used an F-measure metric derived from the precision and recall values for the point-based overall assessment, which is defined as follows}
        
        \begin{equation}
            F-measure = \frac{((1 + \epsilon^2)TP)}{((1+\epsilon^2)TP + \epsilon^2FN + FP)},
        \end{equation}
        
\revisions{ \noindent  where  $\epsilon=1$ is assumed. Moreover, the intersection over union (IoU) is utilized. IoU is defined as the quantity of intersection of the prediction and the groundtruth points divided by the union of them, which is given by}
        
        \begin{equation}
            IoU = \frac{TP}{(TP + FP + FN)}.
        \end{equation}

\revisions{Different metrics are utilized to test the performance of the proposed ground filtering approaches. In particular, Table \ref{table_metric} shows the performance of the two proposed approaches, normal-vector, and voxel-based ground filtering, in terms of accuracy, precision, recall, F-measure, and IoU.  Besides, the four scans are evaluated, Paris, Lille11, Lille12, and Lille2.  According to the five metrics results, the normal-vector approach slightly outperforms the voxel-based filtering.}

\begin{table}[h!]
\scriptsize
\caption{\revisions{Performance of the proposed ground filtering methods on the Paris-Lille-3D dataset for four different scans.} \vspace{-2mm}}
\label{table_metric}

\setlength{\tabcolsep}{4pt}
\begin{center}
	\scriptsize
	\revisions{\begin{tabular}{| c | c || c c  c c |}
 	\hline
 	\hline
 	& & \multicolumn{4}{c|}{Paris-Lille-3D} \\
 	\hline
 	Algorithm & Metric & Paris & Lille11 & Lille12 & Lille2 \\
 	\hline
 	\hline
 	\multirow{3}{*}{Normal-vector} & Accuracy($\%$) & 98.10 & 99.21 & 98.21 & 99.00\\
 	& Precision($\%$) & 99.41 & 98.61 & 98.89 & 98.06 \\
 	& Recall($\%$) & 95.87 & 99.69 & 97.94 & 99.90  \\
 	& F-measure($\%$) & 97.61 & 99.15 & 98.41 & 98.97 \\
 	& IoU($\%$) & 95.34 & 98.32 & 96.88 & 97.97  \\
 	\hline
  	\hline
  	\multirow{3}{*}{Voxel-based} & Accuracy($\%$) & 97.26 & 98.58 & 97.51 & 98.46 \\
	& Precision($\%$) & 95.52 & 97.09 & 97.58 & 96.93  \\
 	& Recall($\%$) & 97.82 & 99.93 & 98.05 & 99.95  \\
 	& F-measure($\%$) & 96.66 & 98.49 & 97.81 & 98.42 \\
 	& IoU($\%$) & 93.53 & 97.03 & 95.73 & 96.89  \\
 	\hline
 	\hline
	\end{tabular}}
	\vspace{-15pt}
\end{center}
\end{table}

\begin{figure}[h!]
    \begin{center}
    \begin{tabular}{c c c c}
    \hspace{-12pt}
    \includegraphics[width=.33\linewidth]{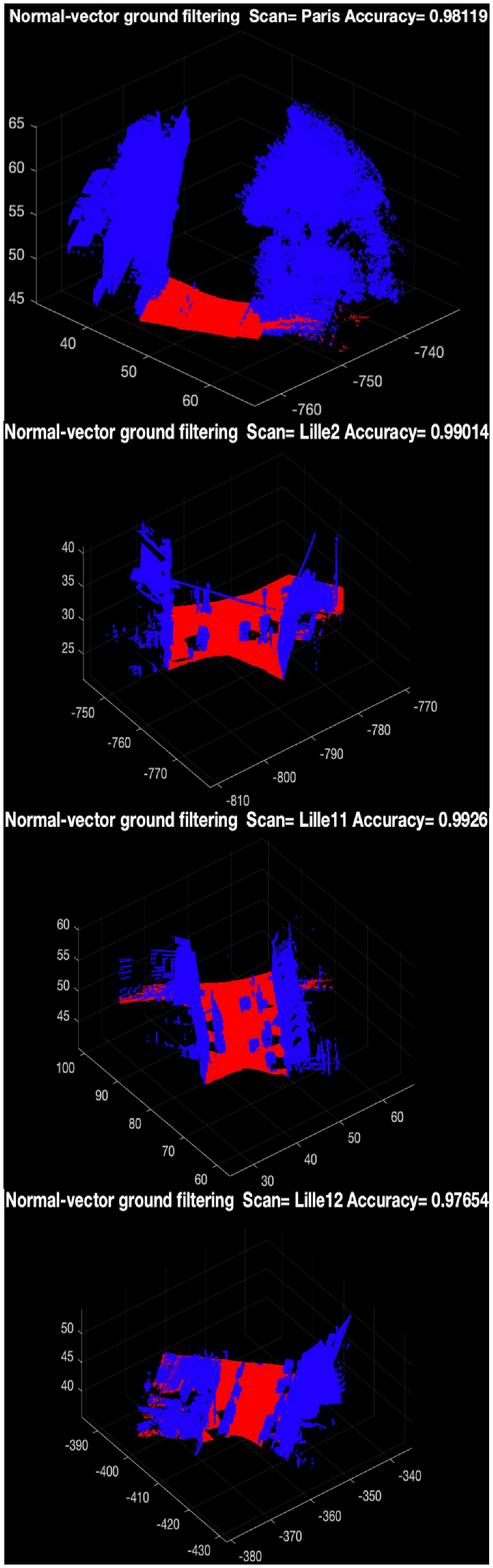}
    &
    \hspace{-12pt}
    \includegraphics[width=.33\linewidth]{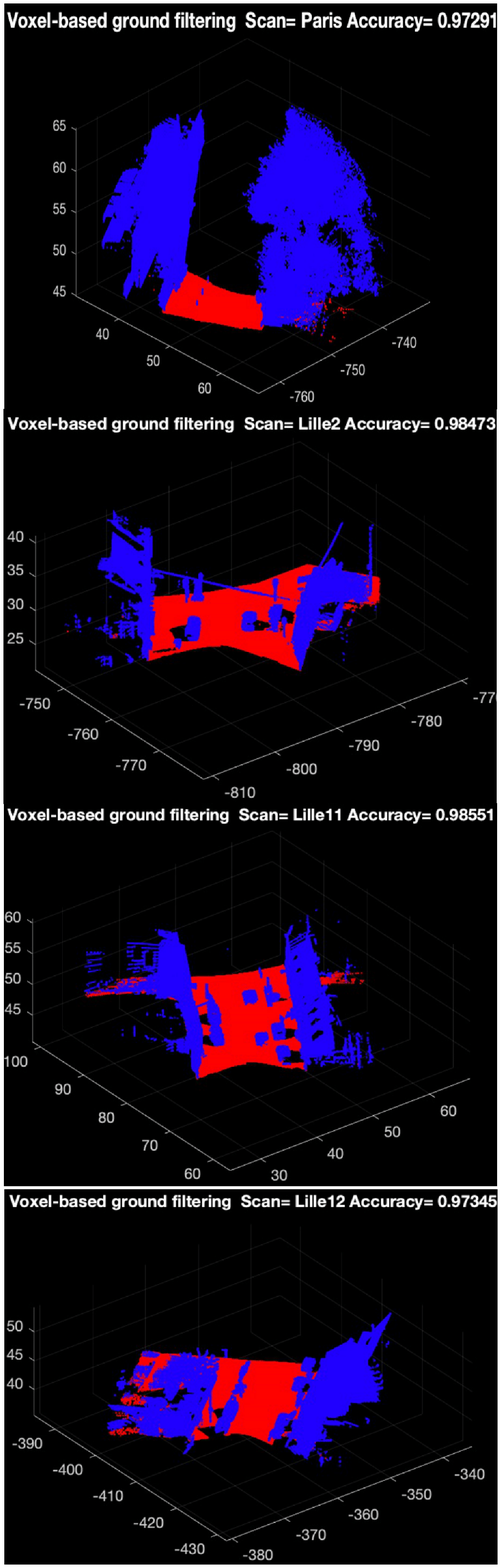}
    &
    \hspace{-12pt}
    \includegraphics[width=.33\linewidth]{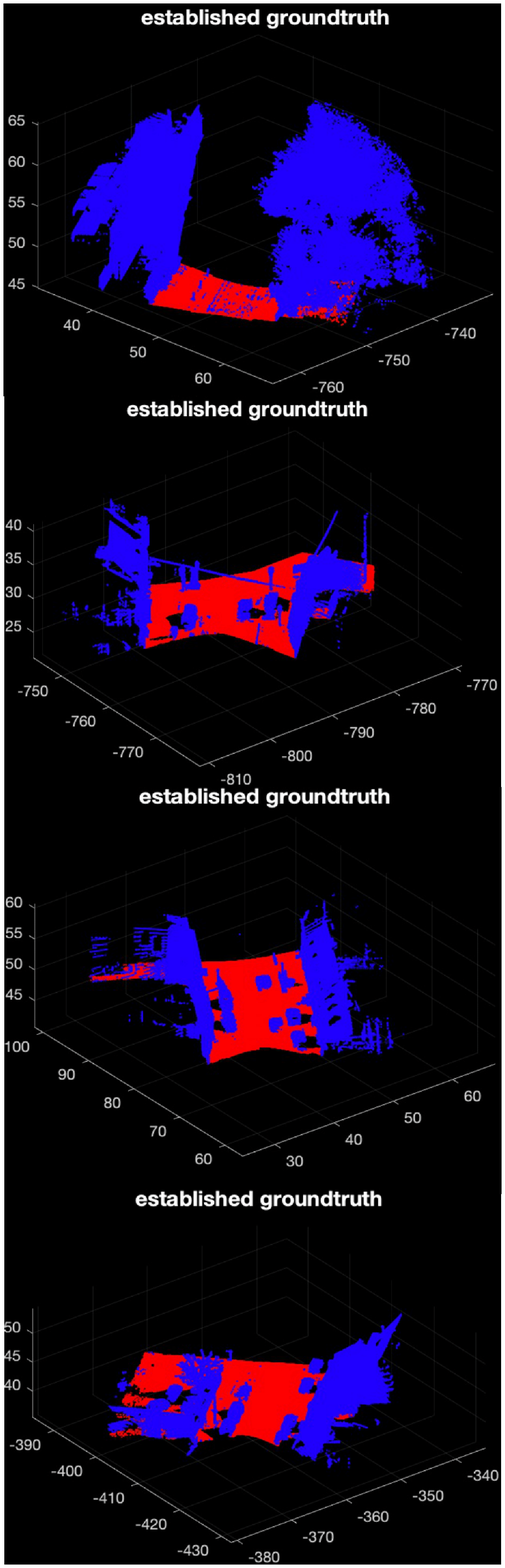}
    \\
    \footnotesize (a) 
    &
    \footnotesize (b)
    &
    \footnotesize (c)
    \end{tabular}\vspace{-8pt}
    \end{center}
    \caption{\revisions{Comparison normal-vector ground filtering, and voxel-based ground filtering using four scans of the Paris-Lille-3D dataset. The first, second, third, and fourth rows correspond to Paris, Lille2, Lille11, and Lille12 scans, respectively. (a) ground filtering using normal-vector approach, (b) ground filtering using voxel-based approach, (c) established groundtruth map.}\vspace{-15pt}}
    \label{fig:paris-lille_ground-filtering}
\end{figure}

\revisions{We evaluate the effectiveness of the proposed ground filtering approaches using four scans of the benchmark Paris-Lille-3D dataset. Figure \ref{fig:paris-lille_ground-filtering} depicts the visual comparison of the normal-vector approach and the voxel-based ground filtering against the groundtruth of the benchmark dataset. In accuracy, the normal-vector outperforms the voxel-based ground filtering. Specifically, the normal-vector obtains accuracy $1\%$ higher than the voxel approach. Although both approaches are comparable in terms of accuracy, the normal-vector approach is computational slower than the voxel approach. Urban scenarios with high density in the point cloud might be adequate for using a voxel-based ground filtering approach.} 

\begin{figure}[h!]
    \begin{center}
    \begin{tabular}{c c c c}
    \hspace{-12pt}
    \includegraphics[width=.33\linewidth]{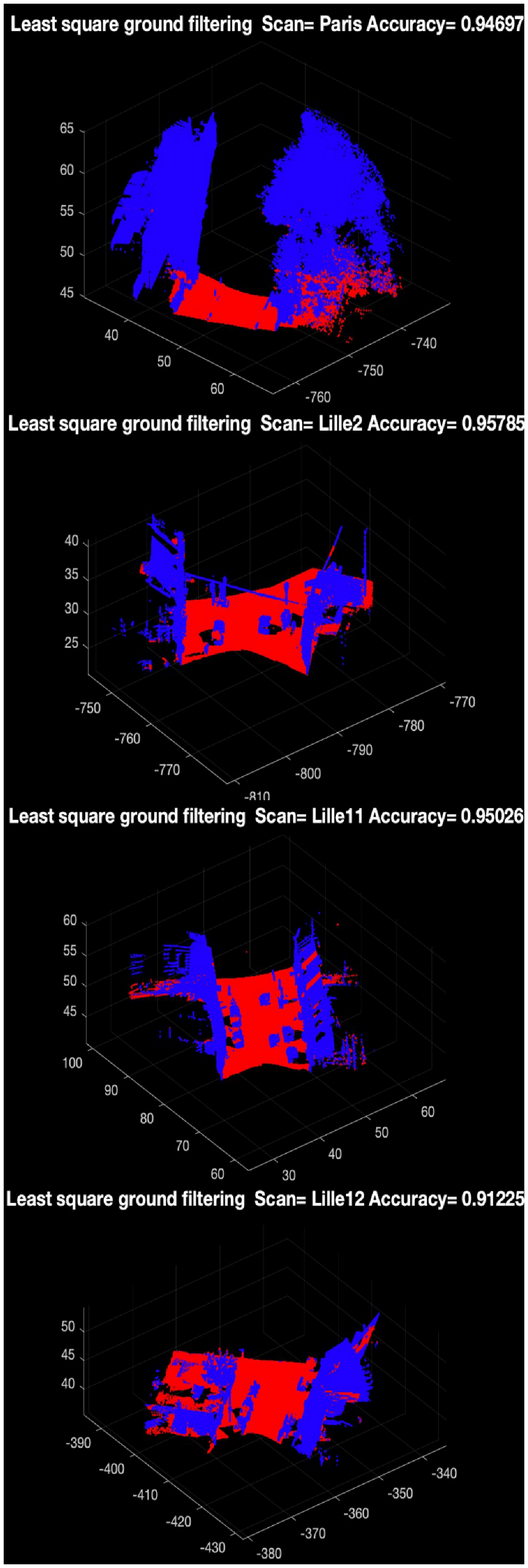}
    &
    \hspace{-12pt}
    \includegraphics[width=.33\linewidth]{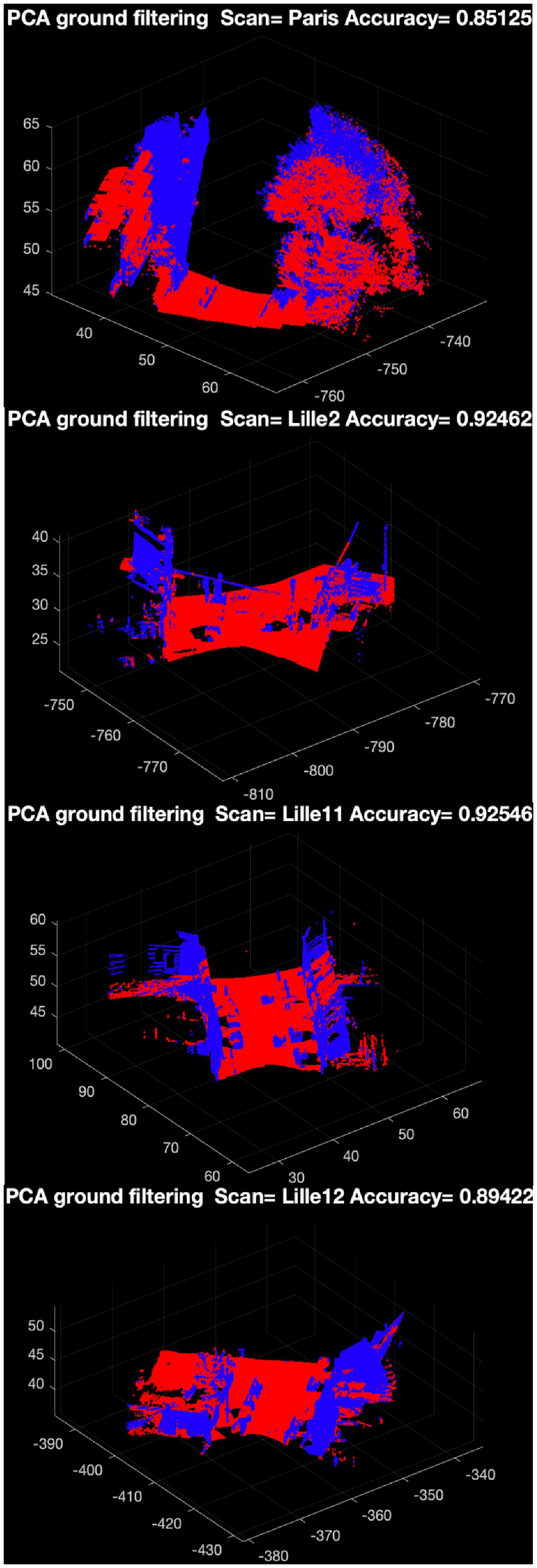}
    &
    \hspace{-12pt}
    \includegraphics[width=.33\linewidth]{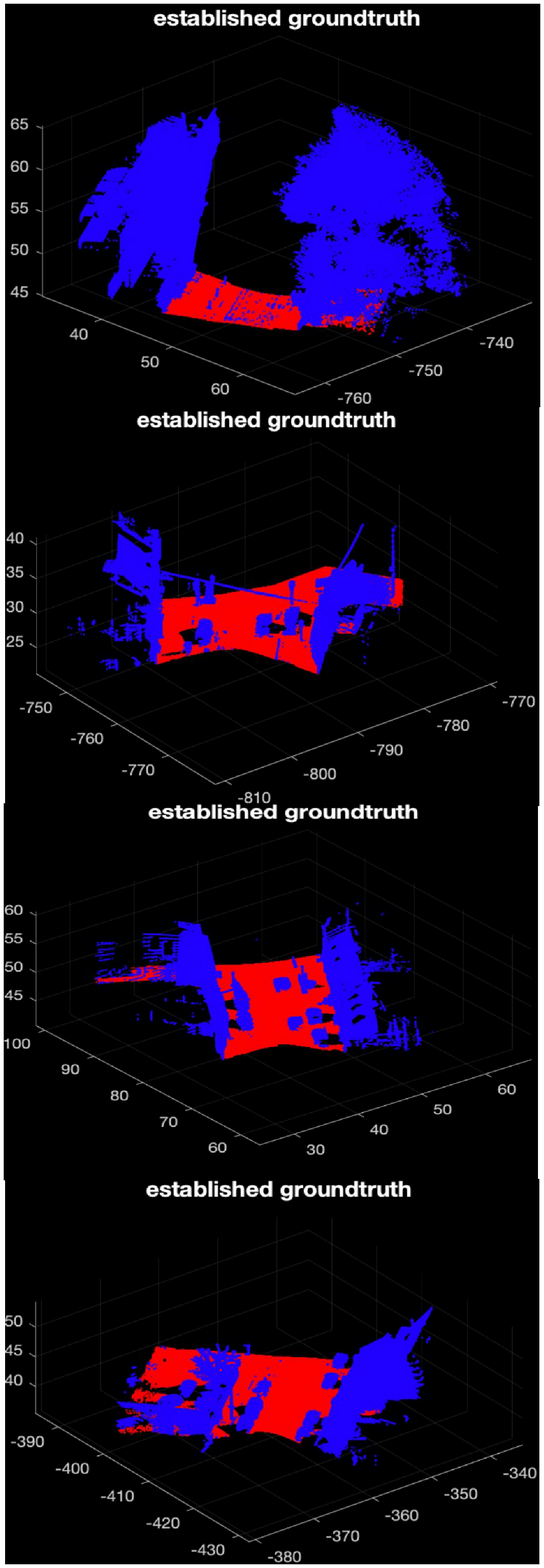}
    \\
    \footnotesize (a) 
    &
    \footnotesize (b)
    &
    \footnotesize (c)
    \end{tabular}\vspace{-8pt}
    \end{center}
    \caption{\revisions{Comparison least square ground filtering, and PCA ground filtering using four scans of the Paris-Lille-3D dataset. The first, second, third, and fourth rows correspond to Paris, Lille2, Lille11, and Lille12 scans, respectively. (a) ground filtering using least square approach, (b) ground filtering using PCA approach, (c) established groundtruth map.}\vspace{-15pt}}
    \label{fig:state-of-art}
\end{figure}

\revisions{Methods of the state-of-art are included to assess the performance of the proposed ground filtering methods. The two methods of the state-of-art are least-square fitting (LS) and principal component analysis (PCA) \cite{10.1145/142920.134011}. In particular, each row of Figure \ref{fig:state-of-art} depicts the ground classification results using one out of four scans of the Paris-Lille-3d dataset. Besides, Figure \ref{fig:state-of-art}(a) shows the ground filtering using least square fitting. Figure \ref{fig:state-of-art}(b) depicts ground filtering using PCA and, Figure \ref{fig:state-of-art}(c) corresponds to the groundtruth map. The accuracy of the LS methods is higher than the PCA for whole scans. In general, LS performs better than PCA. However, the proposed approaches overcome the classification ground filtering of both LS and PCA.}

\subsection{Parameter Sensitivity}

\subsubsection{Computation of Nearest Neighbor}

Table 2 shows the times in minutes of the ground filtering algorithm based on the normal vector, varying the method to calculate the nearest k-neighbors. A GPU is used to implement the KNN algorithm of the CUDA library \cite{4563100}, \cite{5654017} and the kd-tree structuring algorithm of the VTK® library \cite{567752}, \cite{865875}. The average time per section using the KNN CUDA algorithm was 7.0533 minutes compared to the average time per section using the kd-tree data structuring algorithm, which was 5.0705 minutes.  In terms of average time, the kd-tree algorithm was faster compared to the parallelized KNN CUDA algorithm. However, the voxel-based ground filtering algorithm shows an average time of 7.053 seconds for all sections. This is considerably less than the two versions of the ground filtering algorithm based on a normal vector.  It is worth emphasizing that the time measured for the normal-based filtering algorithm with its two variants KNN CUDA and kd-tree VTK, and the voxel-based filtering algorithm include all the steps of the algorithm. The algorithms' lapses based on normal vectors are shown in figure 5.

\begin{table}[!htb]
\centering\footnotesize\caption{Time of the algorithm using KNN CUDA and kd-tree.}
\begin{tabular}{|p{0.8cm}|p{1.0cm}|p{1.8cm}|p{1.8cm}|p{1.8cm}|}\hline
Section & Total points&	Filtering time with kNN CUDA \textbf{(minutes)}&	Filtering time with kd-tree VTK® \textbf{(minutes)}&	Filtering time with voxel \textbf{(seconds)}\\\hline
1 &	4531400	& 6.6090 &	5.9525 & \textbf{6.592} \\
2 &	5330969	& 8.4784 &	7.3477 & \textbf{7.465} \\
3 &	4555752	& 9.5016 &	6.0911 &	\textbf{6.6460} \\
4 &	1269017	& 1.7851 &	1.7526 &	\textbf{3.8650} \\
5 &	5200564	& 11.0656 &	7.0948 &	\textbf{8.3440} \\
6 &	5523518	& 7.3780 &	7.5220 &	\textbf{8.0660} \\
7 &	3098999	& 6.7554 & 4.0597 &	\textbf{4.1880} \\
8 &	3750184	& 9.1994 &	5.0981 &	\textbf{5.8520} \\
9 &	4678069	& 12.6099 &	6.4554 &	\textbf{5.5120} \\
10 &	1047348 & 	2.6767 &	1.3546 &	\textbf{1.5470} \\
11 &	2247984	& 1.5271 &	3.0465 &	\textbf{2.919} \\
\multicolumn{2}{|c|}{Average time} &	7.0533 &	5.0705 &	7.053 \\ \hline
\end{tabular}
\end{table}

\begin{figure}[!htb]
	\centering
	\includegraphics[scale=0.35]{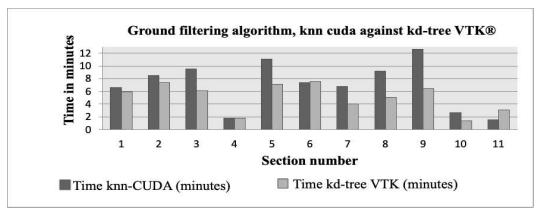}
	\caption{Computation time of the ground filtering algorithm by varying the function that calculates the k- nearest neighbors.}
\end{figure}

Table 3 shows ground filtration results using the normal vector method and the voxel method. $\%\textbf{C}$ shows the percentage of points correctly filtered, while $\%\textbf{E}$ shows the percentage of points incorrectly classified.   The two methods' results are comparable in terms of percentage of points filtered and percentage of error.  The normal vector method presents better results for three sections, while the voxel method presents better results for ten sections.

\begin{table}[!htb]
\centering\footnotesize\caption{Percentage of points correctly and incorrectly filtered applying both methods.}
\begin{tabular}{|p{0.7cm}|p{1.0cm}|p{0.9cm}|p{0.9cm}|p{0.9cm}|p{0.9cm}|}\hline
Section &	Total points&	\multicolumn{2}{c}{Normal vector method} &	\multicolumn{2}{|c|}{Voxel method}\\\hline
& & \%\textbf{C} & \%\textbf{E} & \%\textbf{C} & \%\textbf{E} \\\hline
1&	4531400	&97.42&	2.58&	\textbf{97.63}&	\textbf{2.37}\\
2&	5330699&	92.04&	7.96&	\textbf{98.35}&	\textbf{1.65}\\
3&	4555752&	\textbf{98.13}&	\textbf{1.87}&	97.96&	2.04\\
4&	1474729&	95.25&	4.75&	\textbf{98.43}&	\textbf{1.57}\\
5&	5200564&	94.23&	5.77&	\textbf{95.12}&	\textbf{4.88}\\
6&	5523518&	\textbf{97.43}&	\textbf{2.57}&	97.14&	2.86\\
7&	3098999&	\textbf{98.05}&	\textbf{1.95}&	97.79&	2.21\\
8&	3750184&	98.54&	1.46&	\textbf{98.61}&	\textbf{1.39}\\
9&	4678069&	92.83&	7.17&	\textbf{98.44}&	\textbf{1.56}\\
10&	1047348&	98.64	&1.36&	\textbf{99.01}&	\textbf{0.99}\\
11&	2247981& 	91.77	&8.23&   \textbf{98.73}& \textbf{1.27}\\
\multicolumn{1}{|c|}{Total} &	42063030 &	95.67 &	4.33 & 97.64 & 2.36\\ \hline
\end{tabular}
\end{table}

Table 4 shows the confusion matrix for each of the sections used.  The confusion matrix compares the ground filtration developed with the normal vector method and the voxel method to the ground truth for each one of the sections.  P and N are Positive and Negative, respectively.  The confusion matrix shows that the Normal vector method and the Voxel method present comparable results with respect to the ground truth.

\begin{table}[!htb]
\centering\footnotesize\caption{Confusion matrix for each of the ground filtering methods.}
\begin{tabular}{|p{0.8cm}|p{1.5cm}|p{0.9cm}|p{0.9cm}|p{0.9cm}|p{0.9cm}|}\hline
Section&Groundtruth&	\multicolumn{4}{|c|}{Classification} \\ \hline
& & \multicolumn{2}{|c|}{Normal vector method} &\multicolumn{2}{|c|}{Voxel method}\\\hline
& &P &N &P &N \\\hline
1&	P&	2065224&	13128&	2129823&	68301\\
& 	N&	103842&	2349206&	39243&	2294033 \\
2&	P&	1860416&	11410&	2221623&	36444\\
& 	N&	412685&	3046188&	51478&	3021154\\
3&	P&	2894538&	29504&	2875919&	18907\\
& 	N&	55518&	1576192&	74137&  	1586789\\
4&	P&	468780&	1179&	526872&	12481\\
& 	N&	68812&	935958&	10720&	924656\\
5&	P&	2833947&	266220&	2784639&	170330\\
& 	N&	33937&	2066460&	83245&	2162350\\
6&	P&	3116226&	115567&	3092499&	107531\\
& 	N&	26554&	2265171&	50281&	2273207\\
7&	P&	2339287&	40841&	2314216&	23938\\
& 	N&	19464&	699407&	44535&	716310\\
8&	P&	2391553&	40497&	2378583&	25090\\
&	N&	14095&	1304039&	27065&	1319446\\
9&	P&	2620212&	11550&	2897532&	26545\\
&	N&	323707&	1722600&	46387&	1707605\\
10&	P&	755722&	8025&	758565&	6993\\
&	N&	6197&	277404&	3354&	278436\\
11&	P&	1738240&	5905&	1898857&	10033\\
&	N&	179147&	324689&	18530&	320561\\ \hline
\end{tabular}
\end{table}

\section{Conclusions}
In the experimentation, the average time of all sections for each algorithm was calculated.  The voxel-based ground filtering was found to be faster in terms of average time than the normal vector algorithm, i.e., the average time of the voxel algorithm was 7.053 seconds and was less than those of the ground filtering that uses the search for the nearest k-neighbors. Two algorithms were used to structure data using k-neighbors: the kd-tree version that took 5.0705 minutes and the KNN CUDA version of the algorithm took 7.0533 minutes.  Both algorithms proved to be effective in recognizing ground in urban settings. As part of future studies, it is planned to make a voxel-based implementation of segmentation and classification algorithms for point clouds that uses the proposed ground filtering algorithm during the preliminary stage. Studies that consider non-urban settings are also recommended.

\section{Funding}
Nelson Diaz was supported by Colciencias during his doctoral studies with grant ``727 Doctorados nacionales de 2015". Part of this research project was financed by the Geomatics research group, attached to the Civil Engineering School of the Industrial University of Santander.  Nelson Díaz is supported Universidad de Investigación y Desarrollo (UDI) under internal grant: ``2021-INV-11-BGA Procesamiento de nube de puntos adquiridas con un escáner laser terrestre mediante técnicas de aprendizaje de máquina (machine learning)".


\bibliography{mybibfile}

\end{document}